\documentclass[11pt,a4paper]{article}
\usepackage[hyperref]{acl2019}
\usepackage{times}
\usepackage{latexsym}

\usepackage{url}

\usepackage{adjustbox,multirow}
\usepackage{mathrsfs}
\usepackage{amsmath}
\usepackage{algorithm}
\usepackage{algorithmic}
\usepackage{multirow}
\usepackage{tabularx}
\usepackage{graphicx}
\usepackage{amssymb}

\aclfinalcopy 
\def\aclpaperid{2171} 

\title{Domain Adaptive Dialog Generation via Meta Learning}

\author{Kun Qian \\
  Univeristy of California, Davis \\
  \texttt{kunqian@ucdavis.edu} \\\And
  Zhou Yu \\
  Univeristy of California, Davis \\
  \texttt{joyu@ucdavis.edu} \\}

\date{}

\begin{document}
\maketitle
\begin{abstract}
Domain adaptation is an essential task in dialog system building because there are so many new dialog tasks created for different needs every day. Collecting and annotating training data for these new tasks is costly since it involves real user interactions. We propose a domain adaptive dialog generation method based on meta-learning (DAML). DAML is an end-to-end trainable dialog system model that learns from multiple rich-resource tasks and then adapts to new domains with minimal training samples.
We train a dialog system model using multiple single-domain dialog data with rich-resource by applying the model-agnostic meta-learning algorithm to dialog domain. The model is capable of learning a competitive dialog system on a new domain with only a few training examples in an efficient manner. The two-step gradient updates in DAML enable the model to learn general features across multiple tasks. We evaluate our method on a simulated dialog dataset and achieve state-of-the-art performance, which is generalizable to new tasks.

\end{abstract}

\section{Introduction}
Modern personal assistants, such as Alexa and Siri, are composed of thousands of single-domain task-oriented dialog systems.
Every dialog task is different, due to the specific domain knowledge. An end-to-end trainable dialog system requires thousands of dialogs for training.
However, the availability of the training data is usually limited as real users have to be involved to obtain the training dialogs. Therefore, adapting existing rich-resource data to new domains with limited resource is an essential task in dialog system research. 
Transfer learning~\citep{Caruana:1997:ML:262868.262872, bengio2012deep, cohn1994improving,mo2018personalizing}, few-shot learning \citep{salakhutdinov2012one,li2006one,norouzi2013zero,socher2013zero} and meta-learning \cite{DBLP:journals/corr/FinnAL17} are introduced in solving such data scarcity problem in machine learning. Because every dialog domain is very different from each other, generalize information from rich-resource domains to another low resource domain is difficult. Therefore, only a few studies have tackled domain adaptive end-to-end dialog training methods \cite{DBLP:journals/corr/abs-1805-04803}. We propose DAML based on meta-learning to combine multiple dialog tasks in training, in order to learn general and  transferable information that is applicable to new domains.

\citet{DBLP:journals/corr/abs-1805-04803} introduces action matching, a learning framework that could realize zero-shot dialog generation (ZSDG), based on domain description, in the form of seed response. With limited knowledge of a new domain, the model trained on several rich-resource domains achieves both impressive task completion rate and natural generated response.
Rather than action matching, we propose to use model-agnostic meta-learning (MAML) algorithm~\citep{DBLP:journals/corr/FinnAL17} to perform dialog domain adaptation.
The MAML algorithm tries to build an internal representation of multiple tasks and maximize the sensitivity of the loss function when applied to new tasks, so that small update of parameters could lead to large improvement of new task loss value. This allows our dialog system to adapt to new domain successfully not only with little target domain data but also in a more efficient manner.

The key idea of this paper is utilizing the abundant data in multiple resource domains and finding an initialization that could be accurately and quickly adapted to an unknown new domain with little data. We use the simulated data generated by SimDial~\citep{DBLP:journals/corr/abs-1805-04803}. Specifically, we use three domains: restaurant, weather, and bus information search, as source data and test the meta-learned parameter initialization against the target domain, movie information search.
By modifying Sequicity~\citep{P18-1133}, a seq2seq encoder-decoder network, improving it with a two-stage CopyNet~\citep{Gu2016IncorporatingCM}, we implement the MAML algorithm to achieve an optimal initialization using dialog data from source domains. Then, we fine-tune the initialization towards the target domain with a minimal portion of dialog data using normal gradient descent. Finally, we evaluate the adapted model with testing data also from the target domain.
We outperform the state-of-the-art zero-shot baseline, ZSDG ~\citep{DBLP:journals/corr/abs-1805-04803}, as well as other transfer learning methods \cite{caruana1997multitask}. 
We publish the code on the github\footnote{https://github.com/qbetterk/DAML.git}.

\section{Related Works}
Task-oriented dialog systems are developed to assist users to complete specific tasks, such as booking restaurant or querying weather information. The traditional method to build a dialog system is to train modules separately~\citep{chen2017survey} such as: natural language understanding (NLU)~\citep{Deng2012UseOK,Dauphin2014ZeroShotLA,hashemi2016query}, dialog state tracker~\citep{Henderson2014TheTD}, dialog policy learning~\citep{Cuayhuitl2015StrategicDM, Young2013POMDPBasedSS} and natural language generation (NLG)~\citep{Dhingra2017EndtoEndRL, Wen2015StochasticLG}. \citet{henderson2013deep} introduces the concept of belief tracker that tracks users' requirements and constraints in the dialog across turns.
Recently, more and more works combine all the modules into a seq2seq model for the reason of easier model update. 
\citet{P18-1133} has introduced a new end-to-end dialog system, sequicity, constructed on a two-stage CopyNet~\cite{Gu2016IncorporatingCM}: one for the belief tracker and another one for the response generation.
This model has fewer number of parameters and trains faster than the state-of-the-art baselines while outperforming baselines on two large-scale datasets.

The traditional paradigm in machine learning research is to train a model for a specific task with plenty of annotated data. 
Obviously, it is not reasonable that large amount of data is still required to train a model from scratch if we already have models for similar tasks. Instead, we want to quickly adapt a trained model to a new task with a small amount of new data. 
Dialog adaptation has been explored in various dimensions. \citet{DBLP:journals/corr/abs-1804-10731} introduces an end-to-end dialog system that adapts to user sentiment. \citet{mo2018personalizing} and \citet{Genevay2016TransferLF} also trains a user adaptive dialog systems using transfer learning. 
Recently, effective domain adaptation has been introduced for natural language generation in dialog systems~\citep{Tran2018AdversarialDA, Wen2016MultidomainNN}. Some domain adaptation work has been done on dialog states tracking~\citep{mrkvsic2015multi} and dialog policy learning~\citep{Vlasov2018FewShotGA} as well. However, there is no recent work about domain adaptation for a seq2seq dialog system, except ZSDG~\citet{DBLP:journals/corr/abs-1805-04803}. ZSDG is a zero-shot learning method that adapts action matching to adapt models learned from multiple source domains to a new target domain only using its domain description. Different from ZSDG, we propose to adapt meta-learning to achieve similar domain adaption ability.

\begin{figure*}[htb!]
\centering
\includegraphics[width=13cm]{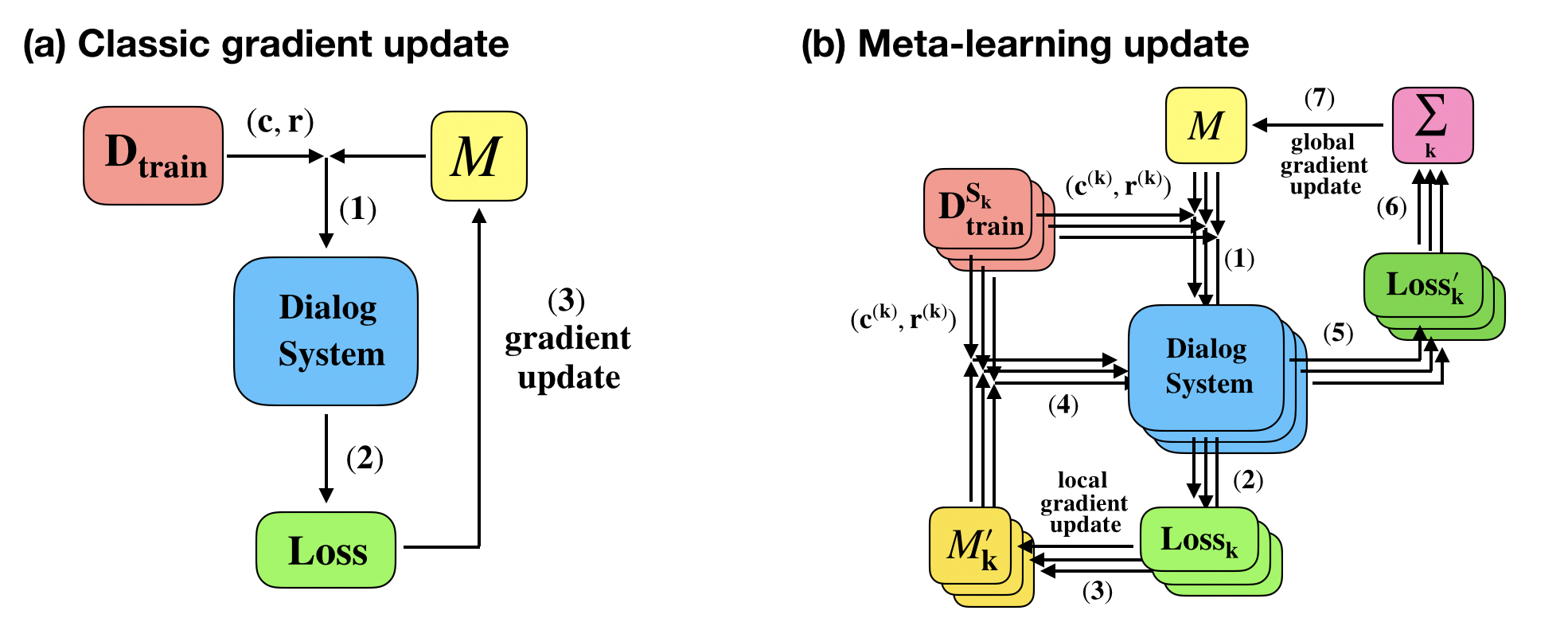}
\caption{(a) shows the classical gradient update steps. (b) shows how we use MAML to update model with gradient descent. The index numbers suggest the processing order of each step.}
\label{fig_model}
\end{figure*}

Meta-learning aims at learning new tasks with few steps and little data based on well-known tasks. One way to realize meta-learning is to learn an optimal initialization that could be adapted to new task accurately and quickly with little data~\citep{vinyals2016matching,snell2017prototypical}. Another way to learn the learning progress is to train a meta-learner to optimize the optimizer of original network for updating parameters~\citep{andrychowicz2016learning, grant2018recasting}. Meta-learning has been applied in various circumstances such as image classification~\citep{Santoro2016MetaLearningWM, DBLP:journals/corr/FinnAL17}, machine translation~\citep{DBLP:journals/corr/abs-1808-08437}, robot manipulation~\cite{DBLP:journals/corr/DuanSCBSA16,DBLP:journals/corr/WangKTSLMBKB16}, etc.  We propose to apply meta-learning algorithm on top of the sequicity model to achieve dialog domain adaptation. Specifically, we chose the recently introduced algorithm, model-agnostic meta-learning(MAML) ~\cite{DBLP:journals/corr/FinnAL17}, because it generalizes across different models. This algorithm is compatible with any model optimized with gradient descent, such as regression, classification and even policy gradient reinforcement learning. Moreover, this algorithm outperforms other state-of-the-art one-shot algorithms for image classification.

\section{Problem Formulation}
\label{sec:problem_formation}
Seq2Seq-based dialog models take the dialog context $c$ as the input and generates a sentence $r$ as the response. Given the abundant data in the $K$ different source domains, we have the training data in each source domain $S_k$, denoted as: 
$$D^{S_k}_{train} =\{(c^{(k)}_n,r^{(k)}_n,S_k), n=1...N\},\ \ k=1...K$$
we also denote the data in the target domain $T$ as:
$$D^{T}_{train} = \{(c^T_n,r^T_n,T), n=1...N'\}$$
where $N'<<N$ and  $N'$ is only $1\%$ of $N$ in our setting.

During the training process, we generate a model $$\mathcal{M}_{source}:C\times S_k\rightarrow R$$
where $C$ is the set of context and $R$ is the set of system responses.

For the adaptation, we fine-tune the model $\mathcal{M}_{source}$ with target domain training data $D^T_{train}$ and obtain a new model $\mathcal{M}_{target}$. Our primary goal is to learn a model that could perform well in the new target domain:
$$\mathcal{M}_{target}:C_{target}\times T\rightarrow R_{target}$$

\section{Proposed Methods}

We first introduce how to combine the MAML algorithm and the sequicity model. As illustrated in the Figure~\ref{fig_model}, the typical gradient descent includes (1) combining training data and initialized model, (2) computing the objective loss and then (3) using the loss to update the model parameters. However, with MAML, there are two gradient update steps.  (1) We first combine the initialized model $\mathcal{M}$ with training data $(c^{(k)}, r^{(k)})$ from each source domain $S_k$ separately. (2) For each dialog domain, we calculate the loss $Loss_k$ and them use it to update every new temporary domain model $\mathcal{M}_k'$. (4) Again we use the data $(c^{(k)}, r^{(k)})$ from each domain and its corresponding temporarily updated domain model $\mathcal{M}_k'$ to calculate a new loss $Loss_k'$ in each domain, (6) then sum all the new domain loss to obtain the final loss. (7) Finally, we use the final loss to update the original model $\mathcal{M}$.

In the following part, we describe the implementation details of the MAML algorithm and the sequicity model separately.
As illustrated in Algorithm \ref{alg1}, sequicity model is used to combine natural language understanding (NLU), dialog managing and response generation in a seq2seq fashion, while meta-learning is a method to adjust  loss function value for better optimization. $\alpha$ and $\beta$ in the algorithm are the learning rate. As mentioned in Section \ref{sec:problem_formation}, $c$ denotes the context and is the input to the model at each turn. In order to use the sequicity model, we format $c$ as $\{B_{t-1},R_{t-1},U_t\}$ at time $t$, where $B_{t-1}$ is the previous belief span at time $t-1$, $R_{t-1}$ is the last system response and $U_t$ is the current user utterance. Sequicity model introduces belief spans to store values of all the informable slots and also record requestable slot names through the history. In this way, rather than put all the history utterances into a RNN to extract context features, we directly deal with the slots stored in the belief span as the representation of all history contexts. The belief span is more accurate and simple to represent the history context and needed to be updated in every turn. The informable and requestable slots are stored in the same span, but with different labels to avoid ambiguity. The context at time $t=1$ contains an empty set as the former belief span $B_0$, and an empty string as the previous system response $R_0$

\begin{algorithm}[t] 
\caption{DAML} 
\label{alg1} 
\begin{algorithmic} 
    \renewcommand{\algorithmicrequire}{\textbf{Input:}}
    \REQUIRE dataset on source domain $D^{S}_{train}$; $\alpha$; $\beta$
    \renewcommand{\algorithmicensure}{\textbf{Output:}} 
    \ENSURE optimal meta-learned model
        \STATE $Randomly\ \ initialize\ \ model \ \  \mathcal{M}$
        \WHILE{$not \ \ done$}
        \FOR{$S_k\in Source\ \ Domain$}
        \STATE Sample data $c^{(k)}$ from $D^{S}_{train}$
        \STATE $\mathcal{M}'_k=\mathcal{M}-\alpha \nabla_{\mathcal{M}}\mathcal{L}_{S_k}(\mathcal{M},c^{(k)})$
        \STATE Evaluate $\mathcal{L}_{S_k}(\mathcal{M}'_k, c^{(k)})$
    \ENDFOR
    
    $\mathcal{M} \leftarrow \mathcal{M}-\beta \nabla_{\mathcal{M}} \sum_{S_k} \mathcal{L}_{S_k}(\mathcal{M}'_k, c^{(k)})$
    \ENDWHILE
    \STATE 
    \renewcommand{\algorithmicrequire}{\textbf{Function}}
    \REQUIRE loss function $\mathcal{L}(\mathcal{M},c)$
        \STATE \textbf{return} \emph{cross-entropy}$(\mathcal{M}(c))$
    \STATE 
    \renewcommand{\algorithmicrequire}{\textbf{Function}}
    \REQUIRE $\mathcal{M}(c^{(k)}=\{B^{(k)}_{t-1},R^{(k)}_{t-1},U^{(k)}_t\})$
  
        \STATE $h = \textrm{Encoder}(B^{(k)}_{t-1},R^{(k)}_{t-1},U^{(k)}_t)$
        \STATE $B_t = \textrm{BspanDecoder}(h)$
        \STATE $R_t = \textrm{ResponseDecoder}(h,B^{(k)}_t,m^{(k)}_t)$

        \STATE \textbf{return} $R_t$  
\end{algorithmic}
\end{algorithm}

The intuition behind the MAML algorithm is that some internal representations are more transferable than others. This suggests that some internal features can be applied to multiple dialog domains rather than a single domain.

Since MAML is compatible with any gradient descent based model, we denote the current generative dialog model as $\mathcal{M}$, which can be randomly initialized. According to the algorithm, for each source domain $S_k$, certain size of training data is sampled. We input the training data $(c^{(k)},r^{(k)})$ into sequicity model and obtain generated system response. 
We adopt cross-entropy as the loss function for all the domains:
$$\mathcal{L}_{S_k}(\mathcal{M},c^{(k)},r^{(k)})=\sum^{|r^{(k)}|}_{j=1} r^{(k)}_j\cdot \log P_{\mathcal{M}}(r^{(k)}_j)$$
For each source domain $S_k$, We use gradient descent to update and get a temporary model.
$$\mathcal{M}'_k \leftarrow \mathcal{M}-\alpha \nabla_{\mathcal{M}}\mathcal{L}_{S_k}(\mathcal{M},c^{(k)},r^{(k)})$$
To be consistent with \cite{DBLP:journals/corr/FinnAL17}, we only update the model for one step. In this way, we have an updated model in each source domain, one step away from $\mathcal{M}$. We may consider multiple steps of gradient update in the future work. Then, we compute the loss based on the updated model with the same training data in each source domain:
$$Loss = \mathcal{L}_{S_k}(\mathcal{M}'_k, c^{(k)},r^{(k)})$$
After this step, we have meta loss value in each domain. We sum up the updated loss value from all source domains as the objective function of meta-learning:
$$\min_{\mathcal{M}} \textrm{Meta-Loss}=\min_{\mathcal{M}} \sum_{S_k} \mathcal{L}_{S_k}(\mathcal{M}'_k, c^{(k)},r^{(k)})$$
Finally, we update the model to minimize the meta objective function:
$$\mathcal{M} \leftarrow \mathcal{M}-\beta \nabla_{\mathcal{M}} \sum_{S_k} \mathcal{L}_{S_k}(\mathcal{M}'_k, c^{(k)},r^{(k)})$$
Unlike common gradient, in MAML, the objective loss we use to update model is not calculated directly from the current model $\mathcal{M}'_k$, but from the temporary model $\mathcal{M}_k'$.
The idea behind this operation is that the loss calculated from the updated model is obviously more sensitive to the changes in original domains, so that we learn more about the common internal representations of all source domains rather than the distinctive features of each domain. Then in the adaptation step, since the basic internal representation has already been captured, the model is sensitive to the unique features of the new domain. As a result, one or a few gradient steps and minimum amount of data are required to optimize the model to the new domain.

\label{maml_sensitive}

The sequicity model is constructed based on a single seq2seq model incorporating copying mechanism and belief span to record dialog states. Given a context $c$ in the form of $\{B_{t-1},R_{t-1},U_t \}$, the belief span $B_t$ at time $t$ is extracted based on the previous belief span $B_{t-1}$ at time $t-1$, the history response $R_{t-1}$ at time $t-1$ and the utterance $U_t$ at time $t$:
$$B_t=\textrm{seq2seq}(B_{t-1},R_{t-1},U_t)$$
Then, we generate system response based on both context and belief span extracted before:
$$R_t=\textrm{seq2seq}(B_{t-1},R_{t-1},U_t|B_t,m_t)$$
$m_t$ is a simple label that helps generate the response. It checks whether or not requested information is available in the database with constraints stored in $B_t$. $m_t$ has three possible values: no match, exact match and multiple match. $m_t=\textrm{\emph{``no match"}}$ denotes that the system cannot find a match in the database given the constraints, then the system would initiate restart the conversation. $m_t=\textrm{\emph{``exact match"}}$ indicates the system successfully retrieves the requested information and completes the task, then the system would end the conversation. $m_t=\textrm{\emph{``multiple matches"}}$ means there are multiple items matches all the constraints, so more constraints are needed to reduce the range of search in the backend database. So the system will then output a question to elicit more information.

\begin{figure}
\centering
\includegraphics[width=6cm]{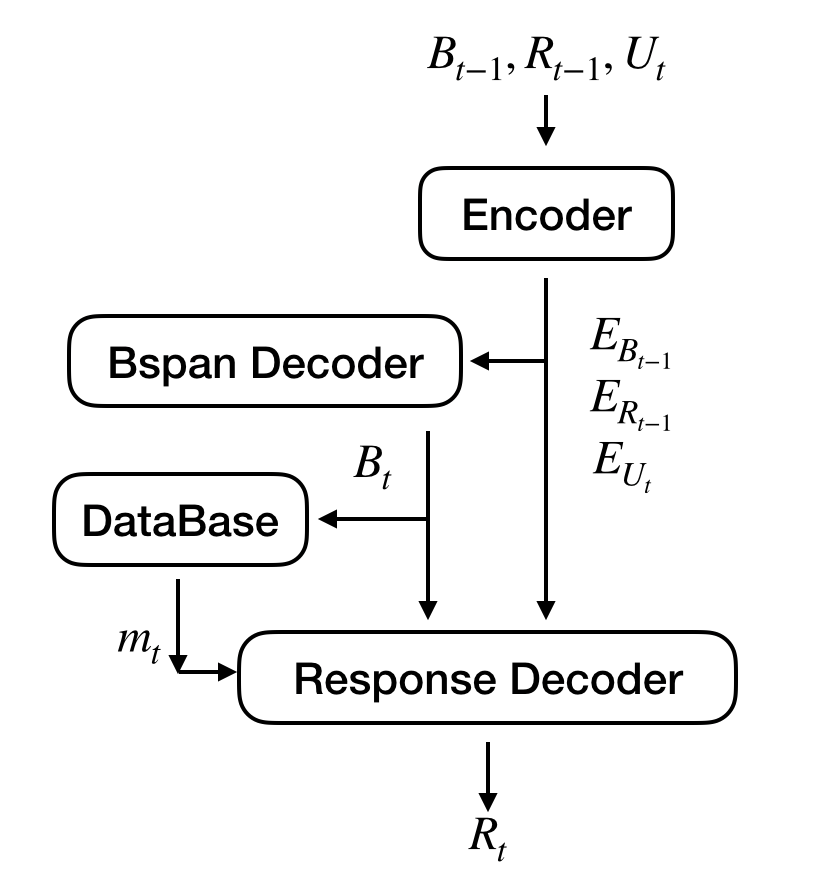}
\caption{Structure of dialog system}
\label{fig_ds}
\end{figure}

The structure is illustrated in Figure \ref{fig_ds} and it is compatible with any seq2seq model. To have a simple architecture, we adopt the basic encoder-decoder structure. Both encoder and decoder employ GRU with attention mechanism.
The response is generated using belief span and utterance at the current time. To simplify the model, we let the belief extractor and response generator share the same encoder. So we reformulate the equations into:
\begin{equation*}
    \begin{split}
        h   & = \textrm{Encoder}(B_{t-1},R_{t-1},U_t) \\
        B_t & = \textrm{BspanDecoder}(h)\\
        R_t & = \textrm{ResponseDecoder}(h,B_t,m_t)
    \end{split}
\end{equation*}
We also need to apply the third attention-based GRU for the response decoding.

Because the response and the utterance usually share some word tokens, the sequicity model also incorporates copy-attention mechanism. Originally, to decode an encoded vector, the model uses softmax to obtain a probability over vocabulary $P^{vocab}(v)$ where $v\in V$. With copy-attention, the decoder not only considers the word generation probability distribution over vocabulary, but also the likelihood of copy the word from input sequence $P^{copy}(v)$ where $v\in V\cup U_t$ and $U_t$ is the current user utterance in the input context $c$. Then the total probability of word $v$ at $i$th token in the output sequence is calculated by summing these two probabilities (normalization is performed after the summation):
$$P_i(v)=(1-g)\cdot P_i^{vocab}(v) + g\cdot P_i^{copy}(v), \ \ v\in V\cup U_t$$ 
The copy probability is calculated similarly in~\citet{Gu2016IncorporatingCM} and is different for belief span decoder and response decoder.

For the belief span decoder, the copy probability is calculated as:

$$P^{copy}_i(v)=\frac{1}{Z} \sum^{|U_t|}_{j:u_j=v} e^{\psi (u_j)} $$
where $Z$ is a normalization factor and $u_j$ is the $j$th word tokens in the utterance $U_t$. We only add the component when $u_j$ is the same as the target word $v$. $\psi(u_j)$ is computed by:
$$\psi(u_j)=\sigma((\textrm{\textbf{h}}_j^{enc})^T\textrm{\textbf{W}})\textrm{\textbf{h}}_j^{dec}$$
where $\textrm{\textbf{h}}_j^{enc}$ is the hidden state in the encoder for the $j$th word as input, $\textrm{\textbf{h}}_j^{dec}$ is the hidden state in the belief span decoder and $\textrm{\textbf{W}}\in \mathbb{R}^{d \times d}$ is the copy-attention weight.

For the response decoder, we apply the copy attention on the recently generated belief span $B_t$ rather than utterance $U_t$:
$$P^{copy}_i(v)=\frac{1}{Z'} \sum^{|B_t|}_{j:b_j=v} e^{\psi (b_j)} $$
$$\psi(b_j)=\sigma((\textrm{\textbf{h}}_j^{dec})^T\textrm{\textbf{W}})\textrm{\textbf{h}}_j^{dec}$$
where both hidden states come from belief span decoder.


\section{Experiment}
We first introduce the dataset and the metrics used to evaluate our models. Then, we describe models evaluated in the experiments and their implementation details.
\subsection{Dataset}
For a fair comparison with the state-of-the-art domain adaptation algorithm, ZSDG ~\cite{DBLP:journals/corr/abs-1805-04803}, we use the dataset, SimDial, which first introduced to evaluate ZSDG. Please refer to Appendix A for an example dialog. There are in total six dialog domains in SimDial: restaurant, weather, bus, movie, restaurant-slot and restaurant-style, where restaurant-slot data has the same slot type and sentence generation templates as the restaurant task but a different slot vocabulary. Similarly, restaurant-style has the same slots but different natural language generation (NLG) templates compared to the restaurant domain.

We choose restaurant, weather and bus as source domains, denoted as following the experiment setting of ZSDG in ~\cite{DBLP:journals/corr/abs-1805-04803}.
For each source domain, we have 900, 100, 500 conversations for training, validation and testing correspondingly, each of which has 9 turns and each utterance has 13 word tokens on average. The rest three domains are for evaluation, which are considered as target domains. The seed response used in ZSDG is a set of system utterances and corresponding labels. To achieve a fair comparison, we use dialog data of the same size for adaptation training. We generate 9 dialogs (1\% of source domain) for each domain's adaptation training, each averagely contains about 8.4 turns. So for each target domain, we assume we have around 76 system response, which is smaller than the 100 seed response, ZSDG used as domain description. For testing, we use 500 dialogs for each target model. Movie is chosen to be the new target domain for evaluation. Because movie has completely different NLG templates and dialog structure, sharing very few common traits with the source domains at the surface level.

\label{ten_random_run}
To avoid any random results in this few-shot learning setting, we report the average of ten random runs for all results. For further exploring the property of the proposed method, we have also generated one dialog for the one-shot experiment, 45 dialogs (5\% of the size in source domain), 90 dialogs (10\% of the size in source domain) study the adaptation efficiency of our methods.
\subsection{Metrics}

There are three main metrics in our experiments: BLEU score, entity F1 score and adapting time. The first two are the most important and persuasive metrics used in \citet{DBLP:journals/corr/FinnAL17} has exhaustively demonstrated the MAML's fast adaptation speed to new tasks. It could even achieve amazing performance with one step of gradient update incorporating with halfcheetah and ant. We would also like to count the number of epochs for adaptation to compare the adaptation speed between our methods and the baseline of transfer learning.
\begin{itemize}
    \item \textbf{BLEU} We use BLEU score~\cite{Papineni2002BleuAM} to evaluate the quality of generated response sentences since generating natural language is also part of the task.
    \item \textbf{Entity F1 Score} For each dialog, we compare the generated belief span and the Oracle one. Since belief span contains all the slots that constraints the response, this score also checks the completeness of tasks.
    \item \textbf{Adapting Time} We count the number of epochs during the adaptation training. We only compare the adaptation with the data of the same size.
\end{itemize}
\subsection{Baseline Models}
To evaluate the effectiveness of our model, we compare DAML with the following two baselines:
\begin{itemize}
    \item \textbf{ZSDG}~\citep{DBLP:journals/corr/abs-1805-04803} is the state-of-the-art dialog domain adaptation model. This model strengthens the LSTM-based encoder-decoder with an action matching mechanism. The model samples 100 labeled utterances as domain description seeds for domain adaptation.
    \item \textbf{Transfer learning} is applied on the sequicity model as the second baseline. We train the basic model by simply mixing all the data from source domains and then following Figure \ref{fig_model} (a) to update the model. We also enlarge the vocabulary with the training data in target domain. Besides, we implement one-shot learning version of this model by only using one target domain dialog for adaptation, as a comparison with the one-shot learning case of DAML.
\end{itemize}

\subsection{Implementation details}
For all experiments, we use the pre-trained GloVe word embedding~\citep{pennington2014glove} with a dimension of 50. We choose the one-layer GRU networks with a hidden size of 50 to construct the encoder and decoder. The model is optimized using Adam~\citep{kingma2014adam} with a learning rate of 0.003. We reduce the learning rate to half if the validation loss increases. We set the batch~\citep{ioffe2015batch} size to 32 and the dropout~\citep{zaremba2014recurrent} rate to 0.5.

\begin{table*}[t]
\centering
\small
\begin{tabular}{c|ccccc}
    \hline
    \textbf{In Domain}  &ZSDG       &Transfer    & DAML      &Transfer-oneshot     & DAML-oneshot \\
    \hline
    BLEU                &70.1       &51.8       &51.8       &51.1       &53.7\\
    Entity F1           &79.9       &88.5       &\textbf{91.4}       &87.6       &91.2\\
    Epoch               &-          &2.7        &1.4        &2.2        &1.0\\
    \hline
    \textbf{Unseen Slot}  &ZSDG       &Transfer    & DAML      &Transfer-oneshot     & DAML-oneshot \\
    \hline
    BLEU                &68.5       &43.3 (46.3)     &41.7 (46.3)         &40.8 (43.9)     &40.0 (41.8)\\
    Entity F1           &74.6       &78.7 (78.5)      &75 (\textbf{79.2})     &70.1 (67.7)        &72.0 (73.0)\\
    Epoch                 &-          &2.6 (2.4)        &4.8 (3.4)    &3.2 (2.6)          &5.0 (3.0)\\
    \hline 
    \textbf{Unseen NLG}  &ZSDG       &Transfer    & DAML      &Transfer-oneshot     & DAML-oneshot \\
    \hline
    BLEU                &70.1       &30.6 (32.4)        &21.5 (26.0)       &20.0 (21.5)    &19.1 (19.1)\\
    Entity F1           &72.9       &82.2 (85.0)       &77.5 (82.4)      &82.8 (86.2)     &69.0 (\textbf{86.4})\\
    Epoch                &-          &3.2 (3.0)        &3.2 (2.1)      &12.3 (20.3)       &4.7 (5.7)\\
    \hline
    \textbf{New Domain}  &ZSDG       &Transfer    & DAML      &Transfer-oneshot     & DAML-oneshot \\
    \hline
    BLEU                &54.6       &30.1      &32.7         &21.5      &22.4\\
    Entity F1           &52.6       &64.0       &\textbf{66.2}      &55.9       &59.5\\
    Epoch                &-          &5.6        &4.5         &14.2      &5.8\\
    \hline

\end{tabular}
\caption{DAML outperforms both ZSDG and transfer learning when given similar target domain data. Even the one-shot DAML method achieves better results than ZSDG. 
Values in parenthesis are the results of the model with an extra step of fine-tuning on the restaurant domain in training. 
``In Domain" uses all three source domains (\textit{restaurant}, \textit{weather} and \textit{bus}),  while ``New Domain" refers to the \textit{movie} domain. ``Unseen Slot" and ``Unseen NLG" correspond to \textit{restaurant-slot} and \textit{restaurant-style} separately.
}
\label{tab_result}
\end{table*}

\section{Results and Analysis}

Table 1 describes all the model performance. We denote testing data from the combination of restaurant, weather and bus domains as ``In Domain" data since they are in the same domains as what we use to train. The data from movie domain is denoted as ``New Domain" as it is unseen in training data. ``Unseen Slot" and ``Unseen NLG" represent restaurant-slot and restaurant-style domains correspondingly.
To keep a fair comparison, both Transfer and DAML use 1\% of source domain data (9 dialogs, in total 76 system responses), which is equal to the seed response that \citet{DBLP:journals/corr/abs-1805-04803} uses. 
We found that both transfer learning and DAML obtain better results than ZSDG. Especially for the ``New Domain", DAML achieves the entity F1 score of $66.2$, $25.8\%$ relative improvement compared with ZSDG. 
As for ``In Domain" testing, DAML also obtains $14.4\%$ improvement beyond ZSDG. However, our method does not get large improvement in the ``Unseen slot" and ``Unseen NLG" domains. 
We notice that these two domains are actually generated from one of the source domain (restaurant domain). So, even though the slots or templates are changed, they should still share some features with the original domain data. If we could take advantage of the original restaurant domain, the result should be improved. Following this intuition, in the ``Unseen slot" domain and the ``Unseen NLG" domain, we first fine-tune the model obtained from DAML with the original restaurant data in training, and then we do further fine-tune with the  adaptation data. The results are further improved and presented in the parenthesis in Table \ref{tab_result}. We see that in most cases, fine-tuning on restaurant data increases both the BLEU score and entity F1 score on the ``Unseen Slot" and ``Unseen NLG" domain.

\citet{DBLP:journals/corr/FinnAL17} emphasizes that meta-learning obtains decent results with extremely small size of data, even in the one-shot cases. To verify this claim, we perform a one-shot version of the DAML training along with one-shot transfer learning by only using one target domain dialog. The result shows that even the one-shot case of DAML outperforms the ZSDG baseline in all cases except ``Unseen slot" in entity F1. For the ``Unseen NLG" domain, the DAML one-shot case even obtains the highest score. 
Considering DAML one-shot also having out-standing performance when adapted to ``In Domain," this suggests that the ``Unseen NLG" domain is relatively close to the ``In Domain." And nearly every model achieves a similarly high score by fine-tuning the model which is already adapted to the ``In Domain" data. Since the score of ``In Domain" is already extremely high, we assume the model have learned the common features well. We also mention in the Sec~\ref{maml_sensitive} that MAML is sensitive to the new knowledge.
Given that the model already learns the common features well ,in the one-shot setting, the model focuses on learning the unique features of the target domain, while the setting with 1\% adaptation data still partially focus on some common features.

And our method shows evident advantage not only with better scores but also with much fewer update steps. We observe in Table~\ref{tab_result}, DAML only needs one epoch to find the optimum when adapting to the ``In Domain." Even for the ``New Domain," DAML only uses $5.8$ epochs on average to converge, which is only $40\%$ of epochs used in transfer learning. The epoch numbers in the Table~\ref{tab_result} are not integers because all the results in our experiment are the average value of results from ten random runs, explained in Sec \ref{ten_random_run}. Therefore, we conclude DAML is more efficient compared with simple transfer learning.

DAML's success mainly comes from three possible reasons. The first is the CopyNet mechanism. The copy model directly copy and output word tokens from the context, contributing to the high entity F1 score. The belief span also helps to improve the performance. With the belief span, we no longer need to extract slots from all the history utterances in each turn. Instead, we only need the previous slots, stored in belief span, that the copy model could directly deal with. This allows us to simplify our framework and improve the performance. Finally, the meta-learning allows our model to learn inner features of the dialog across different domains.

\begin{table}[h]
\centering
\small
\begin{tabular}{l|cc}
    \hline
    \textbf{movie} &Transfer    & DAML \\
    \hline
    Entity F1           &64.0       &\textbf{66.2 }    \\
    BLEU           &30.1       &\textbf{32.7}     \\
    \hline
    \textbf{restaurant} &Transfer    & DAML \\
    \hline
    Entity F1           &80.7       &\textbf{82.1}     \\
    BLEU           &46.1       &\textbf{47.9}     \\
    \hline
    \textbf{bus} &Transfer    & DAML \\
    \hline
    Entity F1           &60.0       &\textbf{61.9}     \\
    BLEU           &32.0       &\textbf{35.9}     \\
    \hline
    \textbf{weather} &Transfer    & DAML \\
    \hline
    Entity F1           &79.1       &\textbf{80.4}     \\
    BLEU           &38.9       &\textbf{43.3}     \\
    \hline

\end{tabular}
\caption{Performance on different dialog domains}
\label{tab_diff_target_domain}
\end{table}

We also change different tasks used in source and target data to validate the robustness of our model. We use the leave-one-out approach to compare the difference between movie, restaurant, bus and weather domains. When we choose one of them as the target domain, we use the other three as the source domains. The size of the dataset (1\% target data for adaptation) and model hyperparameters are keeping the same as the main experiment described above. We observe in the table \ref{tab_diff_target_domain}, the restaurant domain achieves both the highest entity F1 score and the highest BLEU score, which means it is the easiest domain to adapt to. The bus domain receives the lowest entity F1 score and the movie domain holds the second lowest one, as well as the lowest BLEU score. This demonstrates that the movie domain is really a hard domain for adaptation and is worth being chosen as the target domain. Among all combinations, DAML outperforms the transfer learning algorithms in both Entity F1 and BLEU.

\begin{figure}[h]
\centering
\includegraphics[width=8cm]{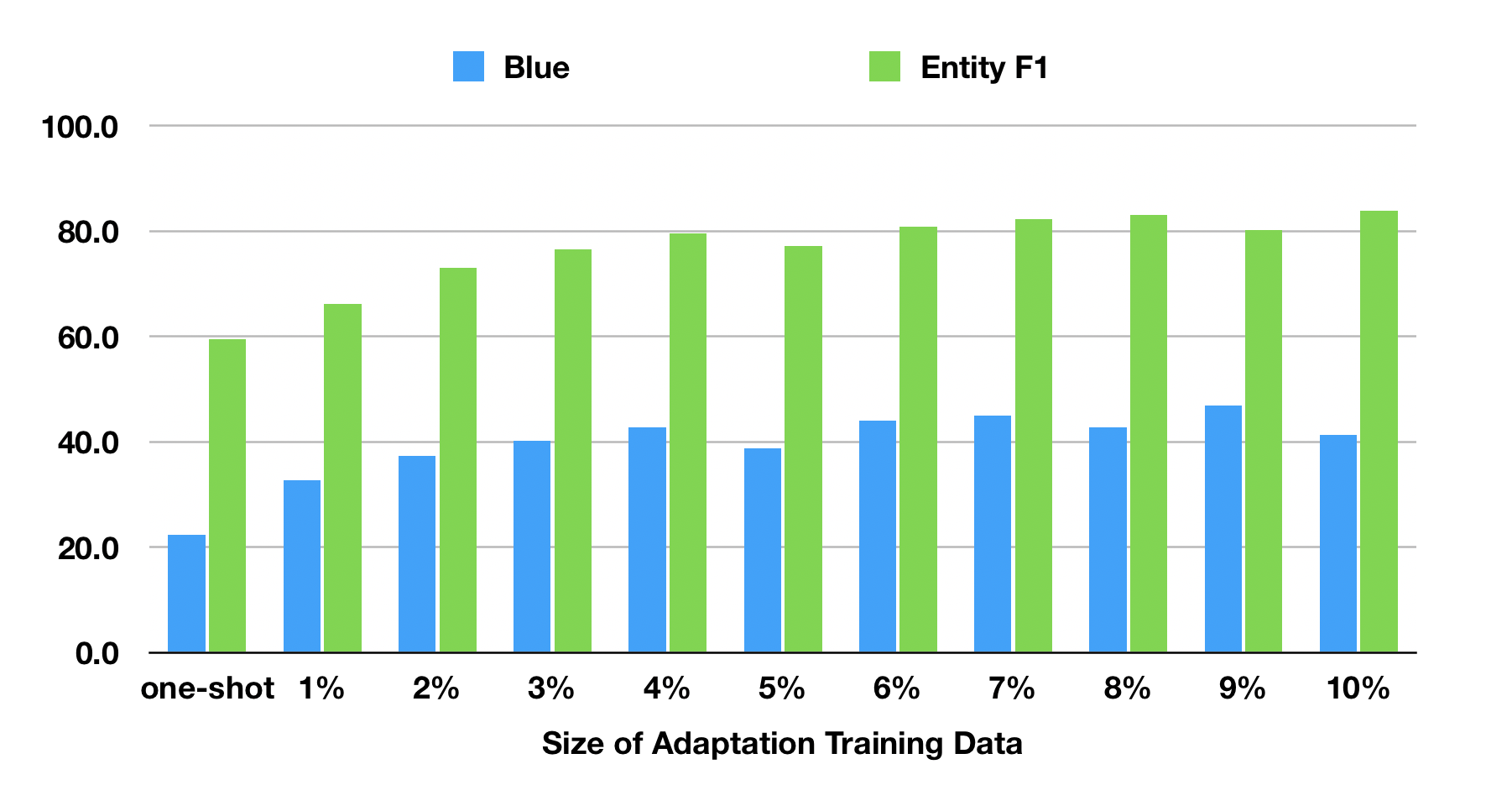}
\caption{The system performance improves when the size of the target data increases. Even the one-shot learning setting achieves decent performance.}
\label{fig_size}
\end{figure}

In addition, we investigate the impact of using different amount of target domain data on system performance. We use the best model trained on restaurant, bus and weather and test on the movie domain. The size of target data varies from one dialog in one-shot learning to 10\% of the data, which is 90 dialogs.  Figure \ref{fig_size} shows the system performance positively correlates with the amount of training data available in the target domain. We observe that both entity F1 and BLEU scores nearly converge when $4\%$ of the data is used. Although $4\%$ is three times the size of the seed response used in \citet{DBLP:journals/corr/abs-1805-04803}, we notice that even the one-shot case of our model outperforms ZSDG in the new domain. This demonstrates our method's capability to achieve good performance with only little target data.

Although the DAML has demonstrated outstanding performance in dialog domain adaptation, it still cannot perfectly adapt to a new domain, especially when there is out of domain words in new domain, denoted as $unk$. If $unk$ lies in the utterance, such as \emph{``system: Movie from what country?"} \emph{``user: Movie from $unk$."} System can hardly extract the needed slot since it does not recognize the surface form of the slot, even if we recognize the $unk$ as the entity. If $unk$ appears in the belief span, when our system uses copy model to generate the new belief span based on the previous one, it is hard to handle the $unk$ token. 

The model also has difficulties in handling complex utterances, especially when a sentence has corrections, such as: \emph{``new request. in 2000-2010. oh no, in 70s."} In this case, our system successfully adds only \emph{70s} to the belief span, mainly because the adverb \emph{in} suggests \emph{70s} is a year. However, the system keeps the original slot year, leading to a \emph{no match} result. Moreover, in the case \emph{``that's wrong. i love western ones."}, our system is confused on what the pronoun \emph{``ones"} refers to. So it does not recognize \emph{``western"} is a dialog slot.

\section{Conclusion and Future Work}
We propose a domain adaptive dialog generation method based on meta-learning(DAML). We also construct an end-to-end trainable dialog system that utilizes a two-step gradient update to obtain models that are more sensitive to new domains.
We evaluate our model on a simulated dataset with multiple independent domains. DAML reaches the state-of-the-art performance in Entity F1 compared with a zero-shot learning method and a transfer learning method. DAML is an effective and robust method for training dialog systems with low-resources. 

The DAML also provides promising potential extension, such as applying DAML on reinforcement learning-based dialog system. We also plan to adapt DAML to multi-domain dialog tasks. 




\bibliographystyle{acl_natbib}
\bibliography{acl2019}



\appendix
\section{SimDial Dataset Example}
\label{sec:appendix}

Table \ref{tab_dialog_eg} is an example dialog in the restaurant domain generated by SimDial. Both the domain and sentence complexity can be adjusted. For each turn, bot the informable and the requestable slots would be recorded and stored in the belief span.

\begin{table*}[hb!]
\centering
\small
\begin{adjustbox}{width=\textwidth}
\begin{tabular}{c|c|c|c|c}
\hline
turn& speaker& utterances & inform slots& request slots\\
\hline
\hline
\multirow{2}{1em}{1}&\textbf{user}&What's up? hmm I am looking for a restaurant.&&\\
                &\textbf{sys}&Which place?&&\\
\hline
\multirow{2}{1em}{2}&\textbf{user}&I uhm yeah I don't care. Oh sorry, Philadelphia.&loc,Philadelphia&\\
                &\textbf{sys}&I believe you said Philadelphia.&&\\
\hline
\multirow{2}{1em}{3}&\textbf{user}&I have more requests. What kind of parking does it have?&loc,Philadelphia;&parking\\
                &\textbf{sys}&The restaurant has no parking. Anything else?&food,Indian&\\
\hline
\multirow{2}{1em}{4}&\textbf{user}&I have more requests. Is hmm ... it closed?&loc,Philadelphia;&opening\\
                &\textbf{sys}&No, It is open right now. What else can I do?&food,Indian&\\
\hline
\multirow{2}{1em}{5}&\textbf{user}&New request. I'm interested in food uhm at Seattle.&loc,Seattle;&\\
                &\textbf{sys}&Do you mean Indian?&food,Indian&\\
\hline
\multirow{2}{1em}{6}&\textbf{user}&Uh-huh.&loc,Seattle;&\\
                &\textbf{sys}&Restaurant 56 is a good choice. What else can I do?
                &food,Indian&\\
\hline
\multirow{2}{1em}{7}&\textbf{user}&Not done yet. What's the average price?&loc,Seattle;&price\\
                &\textbf{sys}&The restaurant serves moderate food.&food,Indian&\\
\hline
\multirow{2}{1em}{8}&\textbf{user}&I have all I need. See you.&loc,Seattle;&\\
                &\textbf{sys}&See you next time.&food,Indian&\\

\hline
\end{tabular}
\end{adjustbox}
\caption{An example dialog generated from SimDial}
\label{tab_dialog_eg}
\end{table*}

\end{document}